\documentclass{article}

\usepackage{arxiv}

\usepackage{amsmath}        
\usepackage[utf8]{inputenc} 
\usepackage[T1]{fontenc}    
\usepackage{hyperref}       
\usepackage{url}            
\usepackage{booktabs}       
\usepackage{amsfonts}       
\usepackage{nicefrac}       
\usepackage{microtype}      
\usepackage{graphicx}
\usepackage[square,numbers]{natbib}
\usepackage{doi}

\title{The Mathematics Behind Spectral Clustering And The Equivalence To PCA}

\date{February 28, 2021}	

\author{
	\href{https://orcid.org/0000-0002-1343-8904}
	{\hspace{1mm}T Shen} \\
	\texttt{ainesmile@gmail.com} \\
}

\renewcommand{\shorttitle}{The Mathematics Behind Spectral Clustering And The Equivalence To PCA}

\hypersetup{
pdftitle={The Mathematics Behind Spectral Clustering And The Equivalence To PCA},
pdfauthor={T Shen},
pdfkeywords={Spectral Clustering, Graph Laplacian, Spectral Embedding, Dimension Reduction, PCA},
}

\begin{document}
\maketitle

\begin{abstract}
	Spectral clustering is a popular algorithm that clusters points
	using the eigenvalues and eigenvectors of Laplacian matrices
	derived from the data. For years, spectral clustering has been
	working mysteriously. This paper explains spectral clustering
	by dividing it into two categories based on whether the graph
	Laplacian is fully connected or not. For a fully connected graph,
	this paper demonstrates the dimension reduction part by offering
	an objective function: the covariance between the
	original data points' similarities and the mapped data points'
	similarities. For a multi-connected graph, this paper proves
	that with a proper $k$, the first $k$ eigenvectors are
	the indicators of the connected components. This paper also
	proves there is an equivalence between spectral embedding
	and PCA.
\end{abstract}


\section{Introduction}


Spectral clustering is a popular algorithm that can be easily solved by standard linear algebra methods.
Despite its simplicity, spectral clustering has been working mysteriously. For years, different papers try to explain
it from different views.
Shi and Malik(2000)\cite{JShi2} use the normalized cuts to measure the total dissimilarity between different groups and the total similarity
within groups. By relaxing indicator vectors to real values, the optimization problem becomes a generalized eigenvalue problem.
However, there is no guarantee on the quality of the relaxed problem's solution compared to the exact solution(von Luxburg, 2007)\cite{ULuxburg6}.
Meilan and Shi(2001)\cite{MMarina4} provide a random walk view of spectral segmentation by interpreting the similarities as edge
flows in a Markov random walk and prove the equivalence between the spectral problem formulated by the normalized cuts method
and the eigenvalues/eigenvectors of the transition matrix of the random walk.
Therefore, the random walk view shares the same problem with normalized cuts.
Saerens, Fouss, Yen, and Dupont(2007)\cite{FFouss1} use Euclidean Commute Time Distance(ECTD) $n(i,j)$ to
measure the average time taken by a random walker for reaching node $j$ when starting from node $i$, and coming back to
node $i$, and make transformations to an n-dimensional Euclidean space preserving ECTD by spectral composition,
and then get the k-dimensional projection by setting the last $n-k$ eigenvectors to zeros.
For a fully connected graph, the commute time distance embedding has a strong connection with spectral embedding.

This paper explains spectral clustering by dividing graph Laplacians into two categories based on whether the graph Laplacian
is fully connected or not. For a fully connected graph, this paper considers spectral clustering as a two-step algorithm that
first finds a low-dimensional representation based on the Laplacian matrix and then applies a classical clustering algorithm to
the low-dimensional representation, such as k-means.
Section 2 focuses on explaining spectral embedding by providing an objective function: the covariance between original data points' similarities and mapped data points' similarities.
Section 3 proves that for a multi-connected graph, with a proper $k$, the first $k$ eigenvectors are the indicators of
the connected components. Section 4 proves there is an equivalence between spectral embedding and PCA.
Section 5 are the conclusions.

\section{Fully Connected Graphs}

Given a set of data points $ X = ( x_1, ... x_n )^{T}, x_i \in R^{m}, i = 1, 2, ..., n$, we can form an undirected graph
$G = (V, E)$ that vertex $v_i$ represents $x_i$. Based on the undirected graph $G$, we can construct a weighted adjacency
graph $W = (w_{ij})_{n \times n} $, where $w_{ij} = w_{ji} \geq 0$. To get $W$, we first construct the adjacency graph
through $\epsilon$-neighborhood, $k$-nearest neighbor or other methods or simply connect all vertices and then put weights
on the connected edges by a similarity function. The most common similarity function is the RBF kernel, where
$w_{ij} = \exp{-\| x_i - x_j \|^2/(2\delta^{2})}$. If $w_{ij} > 0$ for all pair of vertices $v_i$ and $v_j$,
$G$ becomes a fully connected graph that every vertex connects to every other vertex.
 
Given the weighted adjacency matrix, the degree matrix is defined as a diagonal matrix $D = diag(D_{ii})_{n \times n}$, where $ D_{ii} = \sum_{j=1}^{n} w_{ij} $ is the degree of vertex $v_i$. Given $W$ and $D$, we have the unnormalized graph Laplacian

\begin{equation}
L = D - W
\end{equation}

and two common normalized graph Laplacians

\begin{equation}
\begin{split}
& L^{sym} := D^{-\frac{1}{2}} L D^{-\frac{1}{2}} \\
& L^{rw} := D^{-1}L
\end{split}
\end{equation}

where $L^{sym}$ is the symmetric normalized graph and $L^{rw}$ is the random walk normalized graph.

\subsection{The Unnormalized Graph Laplacians}

The absence of objective function for the dimension reduction part is why spectral clustering works mysteriously
for a fully connected graph. A good objective function for dimension reduction should preserve data points' similarities.
Therefore, the covariance between the original data points' similarities and the mapped data
points' similarities should be a good choice.

Let $Y = (y_1, ..., y_n)^{T}$ be the low-dimensional representation, $y_i \in R^{k}$ represents $x_i$, $i = 1, 2, ..., n$,
and $d_{ij}$ be the dissimilarity between $y_i$ and $y_j$, that

\begin{equation}
d_{ij} = \| y_i - y_j \|^2.
\end{equation}

Let the adjacency weight $w_{ij}$ be the similarity between $x_i$ and $x_j$, the covariance between similarities becomes

\begin{equation}
cov(-d, w) = -\frac{1}{2n} \sum_{i=1}^n \sum_{j=1}^n (d_{ij} - \bar{d})(w_{ij} - \bar{w})
\end{equation}

where $w = (w_{11}, w_{12}, ..., w_{nn})$, $d = (d_{11}, d_{12}..., d_{nn})$, and $\bar{d}$, $\bar{w}$ are the means.
Since for any constant $c \neq 0$, we have

\begin{equation}
cov(cd, w) = c * cov(d, w)
\end{equation}

and only the relative information matters for $Y$,
it is natural to constrain the mapped data's means to zero and constant their norms. That is

\begin{equation}
\begin{cases}
\sum_{i=1}^n y_{it} = 0, \ t = 1, ... k \\
\| y_i \|^2 = c_i, i = 1, ..., n
\end{cases}
\end{equation}

where $c_i \neq 0$ is a constant. Therefore, the covariance becomes

\begin{equation}
cov(-d, w) = - \frac{1}{n}  tr(Y^{T}LY) + constant
\end{equation}

and the optimization problem becomes

\begin{equation}\label{optimization_problem}
\begin{aligned}
& \underset{Y}{\text{minimize}}
&& \mathrm{trace}(Y^{T}LY) \\
& \text{subject to}
&& \| y_i \|^{2} = c_i, \ i = 1, 2, ..., n \\ 
&&& \sum_{i=1}^n y_{it} = 0, \ t = 1, 2, ..., k. 
\end{aligned}
\end{equation}

Let $\lambda_0, ..., \lambda_k, ... \lambda_{n-1}$ be the eigenvalues of $L$ sorted in ascending order,
and $f_0, f_1, ..., f_k$ be the first $k+1$ corresponding eigenvectors.
Since the number of zero eigenvalues of a graph Laplacian
equals to the number of connected components, and $L$ is a positive semi-definite fully connected graph Laplacian, we have

\begin{equation}
0 = \lambda_0 < \lambda_1 \leq ... \leq \lambda_{n-1}.
\end{equation}

Let $\vec{1}$ be the n-dimensional vector of all ones and $\vec{0}$ be the n-dimensional vector of all zeros,
since $L \vec{1} =  \vec{0}$, we have

\begin{equation}
\begin{cases}
f_0 = (c_0, ..., c_0)^{T} \\
\vec{1} \cdot f_t = 0, \ t = 1, 2, ..., k
\end{cases}
\end{equation}

where $c_0$ is a constant. As a constant vector, $f_0$ cannot satisfy the constraints in optimization problem
\eqref{optimization_problem} simultaneously. What's more, $L$ is positive semi-definite,
the solution of the optimization problem \eqref{optimization_problem}
becomes the first $k$ non-constant eigenvectors, that is

\begin{equation}
Y^{*} = (f_1, f_2, ... f_k).
\end{equation}

The classical spectral clustering uses the first $k$ eigenvectors, that is

\begin{equation}
Y^{sp} = (f_0, f_1, ..., f_{k-1}).
\end{equation}

As a constant vector, $f_0$ provides no extra information, which means removing $f_0$ makes no difference for 
the clustering result. Therefore, we only need to adjust the number of clusters from $k-1$ to $k$.

\subsection{The Normalized Graph Laplacians}

For the symmetric normalized graph $L^{sym}$, we have 

\begin{equation}
L^{sym}_{ij} = \frac{L_{ij}}{\sqrt{D_{ii} D_{jj}}}.
\end{equation}

Let 

\begin{equation}
z_{i}^{sym} = \frac{y_{i}}{\sqrt{D_{ii}}}
\end{equation}

and 

\begin{equation}
d_{ij}^{sym} = \| z_i^{sym} - z_j^{sym} \|^2.
\end{equation}

Similar to the unnormalized graph, we add constraints to $z_i^{sym}$

\begin{equation}
\begin{cases}
\| z_i^{sym} \| = c_i^{sym} , i = 1, 2, ... n\\
\sum_{i=1}^{n} z_{it} = 0, t = 1, 2, ..., k
\end{cases}
\end{equation}

where $c_i^{sym} \neq 0$ is a constant. Based on the constrains, we have

\begin{equation}
\begin{split}
cov(-d^{sym}, w) & = -\frac{1}{2n} \sum_{i=1}^n \sum_{j=1}^n (d_{ij}^{sym} - \bar{d}^{sym})(w_{ij} - \bar{w})
\\
& = - \frac{1}{n}tr(Y^{T}L^{sym}Y) + constant
\end{split}
\end{equation}

Therefore, the symmetric normalized graph Laplacian optimization problem becomes

\begin{equation}
\begin{aligned}
& \underset{Y}{\text{minimize}}
&& \mathrm{trace}(Y^{T}L^{sym}Y) \\
& \text{subject to}
&& \| z_i^{sym} \| = c_i^{sym} , i = 1, 2, ... n \\
&&& \sum_{i=1}^{n} z_{it}^{sym} = 0, t = 1, 2, ..., k.
\end{aligned}
\end{equation}

Similar to the unnormalized graph Laplacian, the solution becomes the first $k$ non-constant
eigenvectors of $L^{sym}$, that is

\begin{equation}
Y^{sym} = (f_1^{sym}, ... , f_k^{sym})
\end{equation}

and

\begin{equation}
Z^{sym} = \Lambda^{-\frac{1}{2}} Y^{sym}
\end{equation}

where $\Lambda^{-\frac{1}{2}} = diag(\frac{1}{\sqrt{ \lambda_{ii}}})_{n \times n}$ is a diagonal matrix.
It seems that there is a strong connection between the normalized graph Laplacian and the commute time embedding
in \cite{FFouss1}. The random walk normalized graph $L^{rw}$ is similar to the symmetric normalized graph $L^{sym}$,
we only need to set $z_j^{rw} = \frac{y_j}{\sqrt{D_{ii}}}$.

\section{Multi-Connected Graphs}

For a multi-connected graph with $k$ connected components, the weight adjacency matrix will
be a block diagonal matrix after sorting the vertices according to the connected components they belong to.
And the Laplacian matrix $L^{m}$ becomes

\begin{equation}
L^{m} = \begin{bmatrix}
    L_1 &  \\
    & L_2 & \\
    & & \cdots & \\
    & & & L_k
    \end{bmatrix}
\end{equation}

where each subgraph $L_t, t = 1, 2, ..., k$ is a fully connected graph that has one and only one eigenvalue equals to 0.
Let $u_t^{m}$ be the first eigenvector of subgraph $L_t$, $u_t^{m}$ becomes a constant vector with the corresponding eigenvalue equals to zero,
and the first $k$ eigenvectors of $L^{m}$ becomes

\begin{equation}
U^{m} = \begin{bmatrix}
    u_1^{m} &  \\
    & u_2^{m} & \\
    & & \cdots & \\
    & & & u_k^{m}
\end{bmatrix}.
\end{equation}

Therefore, $U^{m}$ becomes the indicator of the connected components.

\section{The Equivalence To PCA}

Principal component analysis(PCA) is a commonly used linear dimension reduction method that attempts to
construct a low-dimensional representation preserving as much variance as possible.
After removing the means, $X^{T}X$ can be recognized as the covariance matrix,
and the $k$-dimensional representation ${Y^{PCA}}^{*}$ becomes

\begin{equation}
{Y^{PCA}}^{*} = X M_k
\end{equation}

where $M_k$ is a $m \times k$ matrix whose columns are the $k$ largest eigenvectors of $X^{T}X$.
That is, the columns of $M_k$ are the eigenvectors of $X^{T}X$ whose corresponding eigenvalues are the $k$ largest,
and the eigenvectors are sorted according to their eigenvalues in descending order.
Denote $\Delta_{k \times k}$ as the diagonal matrix, whose diagonal entries are the sorted $k$ largest eigenvalues of $X^{T}X$, thus

\begin{equation}
X X^{T} {Y^{PCA}}^{*} = X X^{T} X M_{k}  = X  M_{k} \Delta_{k} = {Y^{PCA}}^{*} \Delta_{k}.
\end{equation}

Therefore, ${Y^{PCA}}^{*}$ are the $k$ largest eigenvectors of $G = XX^{T}$.

To make PCA equal to spectral embedding, we need to construct a fully connected graph Laplacian $L^{PCA}$,
whose smallest $k$ non-constant eigenvectors equals to $Y$.
Based on $G$, we choose the cosine similarity as the similarity function,
and to make sure $ w_{ij} > 0 $, we let $w_{ij} = 2 + \cos \theta_{ij} $.
In this way, $L^{PCA}$ becomes a positive semi-definite fully connect graph Laplacian.
After apply standardization to $X$, we have

\begin{equation}
w_{ij} = 2 + x_i^{T}x_j
\end{equation}

and

\begin{equation}
D_{ii} = \sum_{i=1}^n w_{ij} = 2n
\end{equation}

and the Laplacian matrix becomes 

\begin{equation}\label{L_PCA}
L^{PCA} = 2nI - 2H - G
\end{equation}

where $H = (1)_{n \times n}$ is a $ n \times n $ all-ones matrix.
Let $\beta_0,..., \beta_k, ..., \beta_{n-1}$ be the eigenvalues of $L^{PCA}$ sorted in ascending order,
and $u_0, ..., u_k, ..., u_{n-1}$ be the corresponding eigenvectors, we have

\begin{equation}
0 = \beta_0 < \beta_1 ... \leq \beta_k ... \leq \beta_{n-1}
\end{equation}

and the solution of $L^{PCA}$ becomes

\begin{equation}
{Y^{PCA}_{L}}^{*} = (u_1, ..., u_k).
\end{equation}

For $ t = 1, 2, ..., n-1$, since $\vec{1} \cdot u_t = 0$, with equation \eqref{L_PCA}, we have

\begin{equation}
G u_t = (2n - \beta_t) u_t.
\end{equation}

For $\beta_0 = 0$ and $u_0$, since $X$ has zero means and $u_0$ is a constant vector, we have

\begin{equation}
G u_0 = \vec{0}.
\end{equation}

What's more, since $G$ is positive semi-definite, the eigenvalues of $G$ become

\begin{equation}
2n - \beta_1 \geq ... \geq 2n - \beta_k ... \geq 2n - \beta_{n-1} \geq \beta_0 = 0.
\end{equation}

In this way, $u_1, ..., u_k$ become the $k$ largest eigenvectors of $G$
which is the solution of $k$-dimensional PCA, 

\begin{equation}
{Y^{PCA}}^{*} = (u_1, ..., u_k) = {Y^{PCA}_{L}}^{*}.
\end{equation}

\section{Conclusions}

In this paper, we explain the mathematics behind spectral clustering by dividing graph Laplacian
into two categories based on whether the graph is fully connected or not.
For a fully connected graph, we consider spectral clustering as a two-step
algorithm that first reduces dimension and then applies a standard clustering algorithm
to the lower dimensional representation.
We explain the spectral embedding part by offering an objective function which is the
covariance between the original data points' similarities and the mapped data points' similarities.
For a multi-connected graph, we prove that with a proper $k$,
the first $k$ eigenvectors are the indicators of the connected components.

This paper also proves the equivalence between spectral embedding and PCA by setting the cosine similarity
as the similarity function when constructing the weighted adjacency graph.
Since the choice of similarity function is flexible,
the spectral embedding should have more equivalent dimension reduction algorithms.



\begin{thebibliography}{1}

\bibitem{FFouss1}
F.~{Fouss}, A.~{Pirotte}, J.~{Renders}, and M.~{Saerens}.
\newblock Random-walk computation of similarities between nodes of a graph with
  application to collaborative recommendation.
\newblock {\em IEEE Transactions on Knowledge and Data Engineering},
  19(3):355--369, 2007.

\bibitem{JShi2}
{Jianbo Shi} and J.~{Malik}.
\newblock Normalized cuts and image segmentation.
\newblock {\em IEEE Transactions on Pattern Analysis and Machine Intelligence},
  22(8):888--905, 2000.

\bibitem{MBelkin3}
M.~{Belkin} and P.~{Niyogi}.
\newblock Laplacian eigenmaps for dimensionality reduction and data
  representation.
\newblock {\em Neural Computation}, 15(6):1373--1396, 2003.


\bibitem{MMarina4}
M.~Meila and J.~Shi.
\newblock A random walks view of spectral segmentation.
\newblock In T.~S. Richardson and T.~S. Jaakkola, editors, {\em AISTATS}.
  Society for Artificial Intelligence and Statistics, 2001.

\bibitem{SMarco5}
M.~Saerens, F.~Fouss, L.~Yen, and P.~Dupont.
\newblock The principal components analysis of a graph, and its relationships
  to spectral clustering.
\newblock In J.-F. Boulicaut, F.~Esposito, F.~Giannotti, and D.~Pedreschi,
  editors, {\em Machine Learning: ECML 2004}, pages 371--383, Berlin,
  Heidelberg, 2004. Springer Berlin Heidelberg.

\bibitem{ULuxburg6}
U.~von Luxburg.
\newblock A tutorial on spectral clustering.
\newblock {\em CoRR}, abs/0711.0189, 2007.

\bibitem{YWeiss7}
Y.~{Weiss}.
\newblock Segmentation using eigenvectors: a unifying view.
\newblock In {\em Proceedings of the Seventh IEEE International Conference on
  Computer Vision}, volume~2, pages 975--982 vol.2, 1999.

\end{thebibliography}

\end{document}